\documentclass[runningheads]{llncs}
\usepackage[T1]{fontenc}
\usepackage{graphicx,verbatim}

\usepackage{amsmath}
\usepackage{multirow}
\usepackage{soul}
\usepackage{booktabs}
\usepackage{bbding}
\usepackage{makecell}
\usepackage{stfloats}
\usepackage{xcolor}
\usepackage{makecell}
\usepackage{xspace}
\usepackage{url}
\usepackage{wrapfig}
\usepackage{multirow}
\usepackage{makecell}
\usepackage{booktabs, colortbl}
\usepackage{longtable}
\usepackage{hhline}
\usepackage{algpseudocode}
\usepackage{listings}
\usepackage{amssymb}
\begin{document}
\title{WaveFormer: A 3D Transformer with Wavelet-Driven Feature Representation for Efficient Medical Image Segmentation}
%
\author{Md Mahfuz Al Hasan\inst{1} \and
Mahdi Zaman\inst{2} \and
Abdul Jawad\inst{3} \and 
Alberto Santamaria-Pang\inst{4} \and
Ho Hin Lee\inst{4} \and
Ivan Tarapov\inst{4} \and
Kyle See\inst{1} \and
Md Shah Imran\inst{1} \and
Antika Roy\inst{1} \and
Yaser Pourmohammadi Fallah\inst{2} \and
Navid Asadizanjani\inst{1} \and
Reza Forghani\inst{1, 5}}
\authorrunning{F. Author et al.}
%
\institute{University of Florida \and
University of Central Florida \and
University of California Santa Cruz \and
Microsoft Healthcare AI \and
AdventHealth}


\maketitle              
\begin{abstract}

Transformer-based architectures have advanced medical image analysis by effectively modeling long-range dependencies, yet they often struggle in 3D settings due to substantial memory overhead and insufficient capture of fine-grained local features. We address these limitations with WaveFormer, a novel 3D-transformer that: i) leverages the fundamental frequency-domain properties of features for contextual representation, and ii) is inspired by the top-down mechanism of the human visual recognition system, making it a biologically motivated architecture. By employing discrete wavelet transformations (DWT) at multiple scales, WaveFormer preserves both global context and high-frequency details while replacing heavy upsampling layers with efficient wavelet-based summarization and reconstruction. This significantly reduces the number of parameters, which is critical for real-world deployment where computational resources and training times are constrained. Furthermore, the model is generic and easily adaptable to diverse applications. Evaluations on BraTS2023, FLARE2021, and KiTS2023 demonstrate performance on par with state-of-the-art methods while offering substantially lower computational complexity.


\end{abstract}
%
%
%

\section{Introduction}
Medical image segmentation is fundamental to clinical applications such as tumor delineation, organ localization, and surgical planning. Deep learning-based approaches, particularly convolutional neural networks (CNNs), have demonstrated significant success by hierarchically extracting features. However, their limited receptive fields hinder the capture of long-range dependencies, a critical shortcoming in 3D applications where spatial context across distant slices is essential. Vision transformers (ViTs) overcome this limitation by employing self-attention to model global dependencies; yet, their application to 3D volumes is often constrained by substantial memory overhead and computational inefficiency. Hierarchical transformers partially address these issues by restricting self-attention to local windows, but they struggle to capture the fine-grained details necessary for precise volumetric segmentation~\cite{wang2022anti,bai2022improving}.

Hybrid CNN-transformer architectures have thus gained popularity by combining the strengths of both paradigms to preserve local detail through CNN modules while harnessing transformers for global reasoning~\cite{wenxuan2021transbts,chen2021transunet,zhou2021nnformer,heidari2023hiformer}. Although these models enhance performance, they often rely on bulky encoders or complex attention-based decoders—such as UNETR-style architectures~\cite{hatamizadeh2021swin,hatamizadeh2022unetr}, multi-stage pyramids~\cite{azad2023laplacian}, or large transformer blocks~\cite{heidari2023hiformer}—which lead to excessive parameter counts and slower inference times. These pitfalls make current approaches impractical for deployment in resource-constrained clinical environments, where efficiency and scalability are paramount.

Moreover, neuroscientific evidence suggests that human visual processing involves not only a bottom-up mechanism but also a top-down pathway \footnote{In the top-down mechanism, low-spatial frequencies are rapidly projected onto the prefrontal cortex to form abstract object representation, which is then back-projected into the temporal cortex for integration with the bottom-up pathway.~\cite{bar2003cortical,ullman1995sequence}}, where coarse, low-spatial frequency information rapidly reaches higher cortical areas to form an abstract representation before merging with local details~\cite{bar2003cortical,ullman1995sequence}. Motivated by these biologically plausible insights and the need for more efficient transformer designs in 3D, we propose \textbf{WaveFormer}. Our approach leverages the frequency-domain properties of volumetric data by applying the discrete wavelet transform (DWT)~\cite{shensa1992discrete} to partition feature maps into low-frequency (global) and high-frequency (local detail) sub-bands. In doing so, WaveFormer explicitly addresses two critical challenges in 3D segmentation: (1) reducing the heavy computational load of global attention and (2) preserving the detailed structures vital for accurate boundary delineation.

Our contributions are threefold: \textbf{1. Frequency-Domain Representation Learning:} We compute the bulk of self-attention on the low-frequency sub-bands, significantly reducing the token count while retaining global context, with parallel streams preserving the high-frequency details essential for accurate segmentation. 
\textbf{2. Efficient Frequency-Guided Decoder:} Instead of relying on conventional upsampling, we adopt an inverse DWT (IDWT) mechanism to reconstruct segmentation masks from high-frequency components. This approach not only reduces parameter overhead but also enables real-time volumetric inference. 
\textbf{3. Enhanced Local-Global Context Aggregation:} By integrating frequency-domain cues at multiple scales, WaveFormer emulates the top-down processing route of the human visual cortex, effectively fusing coarse global representations with local feature streams to improve 3D segmentation performance.

We evaluate WaveFormer on three major 3D medical benchmarks—BraTS2023, FLARE2021, and KiTS2023. Our experiments demonstrate that WaveFormer achieves competitive or superior accuracy relative to current state-of-the-art methods~\cite{wenxuan2021transbts,chen2021transunet,zhou2021nnformer,heidari2023hiformer}, while substantially lowering model complexity and inference times.

\section{WaveFormer}\label{sec:waveformer}
\begin{figure}[t]
\includegraphics[width=\textwidth]{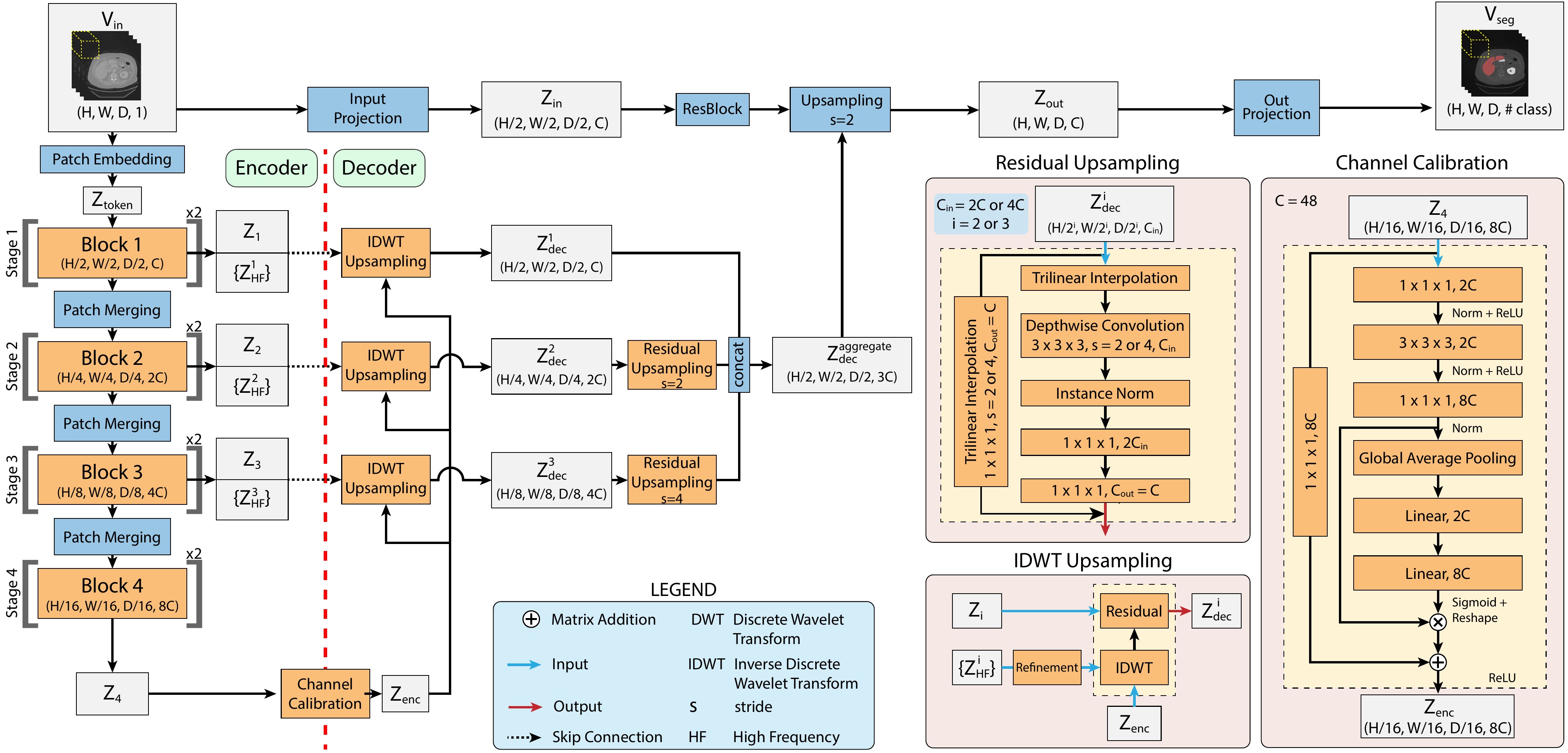}
\caption{The overall architecture of the proposed WaveFormer. The block details are provided in Figure \ref{fig:block}. } \label{fig:network}
\end{figure}

WaveFormer is a hierarchical transformer-based framework designed to address the dual challenges of efficient global context modeling and fine-grained feature preservation while reducing the number of network parameters. As illustrated in Figure~\ref{fig:network}, it integrates two central design principles:

\textbf{Efficient Global Context Modeling:} WaveFormer reduces computational overhead by applying a discrete wavelet transform (DWT) to extract low-frequency sub-bands, enabling self-attention on a compact representation while preserving essential contextual information. \par
\textbf{Detail-Preserving Reconstruction:} High-resolution segmentation masks are progressively reconstructed using IDWT, which reintegrates high-frequency sub-bands to recover fine-grained structural details at each decoding stage.

In addition to these wavelet-based operations, a squeeze-and-excitation module~\cite{hu2018squeeze} serves as a bottleneck for channel calibration, selectively enhancing the most relevant feature channels before the final segmentation output. This mechanism further refines the network’s representational capacity without incurring a substantial computational burden. 

\subsection{Learning on Compact Representations}\label{subsec:dwt}
Recent evidence suggests that self-attention in large-scale transformer models primarily targets low-frequency components~\cite{wang2022vtc}, acting in effect as a low-pass filter. WaveFormer capitalizes on this by employing the discrete wavelet transform (DWT) to decompose volumetric features into multiple resolution sub-bands, capturing coarse global structures in the low-frequency approximation while preserving crucial fine details in the high-frequency components. By structuring features into discrete frequency bands, the model can attend on a computationally compact low-frequency representation—reducing overhead—yet retains high-frequency information essential for precise boundary delineation.

Unlike traditional decoder-heavy architectures that rely on numerous learnable parameters to progressively recover spatial detail, WaveFormer leverages IDWT for upsampling. This approach reintegrates high-frequency components back into the decoded representation, ensuring that the final segmentation output retains both coarse global context and detailed local structures. The multi-resolution capability of wavelets thus offers two key advantages: (i) fewer tokens for global attention, which reduces computational cost, and (ii) efficient reconstruction of fine-scale features, minimizing the need for parameter-intensive decoder blocks. This frequency-centered paradigm opens new possibilities for optimizing transformer-based models, particularly in 3D medical imaging, where multi-scale information can be selectively emphasized to enhance performance and efficiency.

\subsection{Encoder}\label{subsec:encoder}
Given a labeled set of 3D images 
\(
    I_{3D} = \{\,(X_i, Y_i)\,\}_{i=1}^N,
\)
we randomly crop sub-volumes 
\(
    V_i\in \mathbb{R}^{H\times W\times D\times P}
\)
and feed them into the encoder (e.g., \(\!H=W=D=96\) for FLARE). A simple convolution-based patch embedding reduces the input resolution by a factor of two in each spatial dimension, producing initial tokens 
\(
    Z_{\text{token}} \in \mathbb{R}^{\frac{H}{2}\times \frac{W}{2}\times \frac{D}{2}\times C}
\)
with 
\(
    C=48.
\)
These tokens pass through four sequential encoder stages, each comprising two wavelet-attention blocks (Figure~\ref{fig:network}).

\begin{figure}[t]
    \centering
    \includegraphics[width=\textwidth]{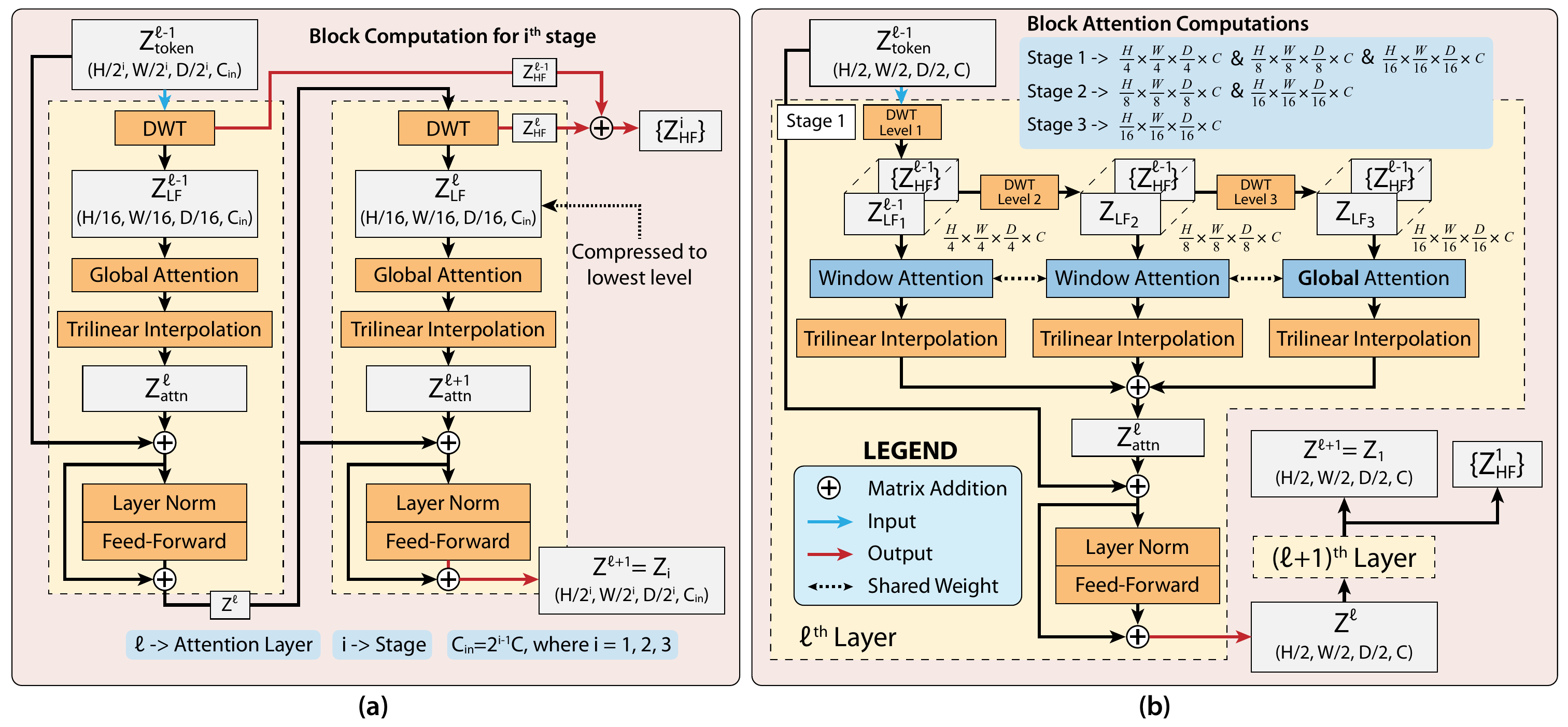}
    \caption{(a) Block architecture of the proposed network for the \(i\)-th encoder stage. Attention is computed solely on the approximation coefficients obtained from multi-level DWT. The HF components extracted at each attention layer are combined and passed to the decoder along with the final stage output. (b) Multi-scale attention design for encoder Stage~1, where attention is computed at each resolution level of the DWT. A fixed window size matching the lowest resolution (third DWT level) enables both global and local context capture in a single attention layer.}
    \label{fig:block}
\end{figure}

\textit{Wavelet-Attention Block.}
In each block, the token feature map undergoes a multi-level discrete wavelet transform (DWT) to separate low-frequency (LF) approximation coefficients \(\,{z}^{l}_{LF}\,\) from high-frequency (HF) detail coefficients \(\,\{z^{l}_{HF}\}\,\). As shown in Eq.~\eqref{eq:attn_blk}, self-attention (MSA) is computed only on the LF approximation, reducing the computational burden while preserving essential global context. The attention output is then upsampled to match the original spatial resolution of the input tokens to that block, processed through a feed-forward network, and forwarded to the next layer as depicted in Figure \ref{fig:block}(a). Symbolically, the wavelet-attention operation in two consecutive layers \(l\) and \(l+1\) can be written as:

\begin{equation}
\begin{array}{ll}
\mathbf{l}^{\text{th}}\ \mathbf{Layer}  & \quad  \mathbf{l+1}^{\text{th}}\ \mathbf{Layer}\\
{z}^{l-1}_{LF}, \{z^{l-1}_{HF}\} = \text{DWT}({z}^{l-1}, \text{level}=m), & \quad {z}^{l}_{LF}, \{z^{l}_{HF}\} = \text{DWT}({z}^{l}, \text{level}=m), \\[6pt]
\hat{z}^{l}_{\text{comp}}=\text{MSA}(\text{LN}({z}^{l-1}_{LF})), & \quad \hat{z}^{l+1}_{\text{comp}}=\text{MSA}(\text{LN}({z}^{l}_{LF})), \\[6pt]
\hat{z}^{l} = \text{Upsample}(\hat{z}^{l}_{\text{comp}})+{z}^{l-1}, & \quad \hat{z}^{l+1} = \text{Upsample}(\hat{z}^{l+1}_{\text{comp}})+{z}^{l}, \\[6pt]
{z}^{l}=\text{MLP}(\text{LN}(\hat{z}^{l}))+\hat{z}^{l}, & \quad {z}^{l+1}=\text{MLP}(\text{LN}(\hat{z}^{l+1}))+\hat{z}^{l+1}, \\ 
\end{array}
\label{eq:attn_blk}
\end{equation}

where \(\,{z}^{l}_{LF}\,\) and \(\,\{z^{l}_{HF}\}\,\) are the LF and HF components from an \(m\)-level DWT in the \(l\)-th layer. \(\,\hat{z}^{l}_{\text{comp}}\)\ represents the attention output on the LF approximation, and \(\,{z}^{l+1}, \{z^{l}_{HF}\}\,\) are the final outputs from an encoder stage.

\textit{Hierarchical Encoding.}
Following the strategy in \cite{he2023swinunetr}, each encoder stage concludes with a patch merging operation that downsamples the feature map by a further factor of two and increases the channel dimension accordingly. Thus, Stage 1 operates on tokens of size 
\(
    \frac{H}{2}\times \frac{W}{2}\times \frac{D}{2}\times C
\),
Stage 2 on
\(
    \frac{H}{4}\times \frac{W}{4}\times \frac{D}{4}\times 2C,
\)
Stage 3 on
\(
    \frac{H}{8}\times \frac{W}{8}\times \frac{D}{8}\times 4C,
\)
and Stage 4 on
\(
    \frac{H}{16}\times \frac{W}{16}\times \frac{D}{16}\times 8C.
\)
Each stage’s output tokens and HF details \(\,\{z^{l}_{HF}\}\,\) are relayed to the decoder. In Stage 4, DWT is omitted because the tokens are already at the lowest spatial resolution.

The network is designed so that, within each wavelet-attention block, the features undergo multi-level DWT with decreasing decomposition levels \(m\) at progressively deeper stages. For instance, Stage~1 applies DWT with \(m=3\) on tokens of resolution 
\(
    \frac{H}{2}\times \frac{W}{2}\times \frac{D}{2}
\)
(i.e., down to 
\(
    \frac{H}{16}\times \frac{W}{16}\times \frac{D}{16}
\))
before performing attention, while Stages 2 and 3 use \(m=2\) and \(m=1\), respectively. This progressive reduction preserves hierarchical representations of global context while controlling computational complexity. 

\textit{Multilevel Attention.}
Utilizing the multi-scale approximation feature from DWT in each layer, we extended our network to compute multi-resolution attention as depicted in Figure \ref{fig:block}(b). For stage 1, attention is computed on each DWT decomposed feature on \(\frac{H}{4}\times \frac{W}{4}\times \frac{D}{4}\), \(\frac{H}{8}\times \frac{W}{8}\times \frac{D}{8}\) and \(\frac{H}{16}\times\frac{W}{16}\times\frac{D}{16}\) scales, respectively. Window attention with the window size matching the lowest-scale feature resolution ($3^{rd}$ level in stage 1) enables capturing contextual relations in both local and global regions within each attention layer, leading to more holistic representation learning. The procedure similarly continues in the $2^{nd}$ and $3^{rd}$ stage encoder blocks with 2 scales and 1 scale attention computations, respectively. 

\subsection{Decoder}\label{subsec:decoder}
At the end of the encoder, the hidden dimension feature map $z_4$ is calibrated using a squeeze-and-excitation module~\cite{hu2018squeeze} to produce a refined embedding $z_{enc}$. $z_{enc}$ is subsequently input into an IDWT upsampling path along with intermediate encoder features $z_i$ and the associated high-frequency (HF) coefficients $\{z^{i}_{HF}\}$ where $i=1,2,3$. A lightweight refinement block suppresses noise in the high-frequency details, and the IDWT reconstructs an upsampled representation that merges with the corresponding skip connection $z_i$ to generate decoder feature map $z^{i}_{dec}$ of dimensions $\tfrac{H}{8}\times \tfrac{W}{8}\times \tfrac{D}{8}\times 4C$, $\tfrac{H}{4}\times \tfrac{W}{4}\times \tfrac{D}{4}\times 2C$, $\tfrac{H}{2}\times \tfrac{W}{2}\times \tfrac{D}{2}\times C$ at stages 3, 2, and 1, respectively. The outputs from stages 3 and 2 are further upsampled using a residual module~\cite{roy2023mednext} and concatenated with the Stage 1 output to form an aggregated feature $z_{\text{dec}}^{\text{aggregate}}$. Finally, fusing $z_{\text{dec}}^{\text{aggregate}}$ with the original patch embedding $x_{in}$ and applying a projection layer produces the final decoded tensor $Z_{\text{out}} \in \mathbb{R}^{H \times W \times D \times C}$ and the segmentation mask $V_{\text{pred}} \in \mathbb{R}^{H \times W \times D \times \text{class}}$.

\section{Experiments} \label{sec:experiments}

\subsection{Datasets} \label{subsec:datasets}
We assessed WaveFormer along with comparable baseline models on three tasks utilizing three publicly available datasets: FLARE2021 \cite{ma2022fast} contains $361$ abdomen CT volumes for multi-organ segmentation. Each volume includes four segmentation targets (spleen, kidney, liver, and pancreas). KiTS2023 \cite{myronenko2023automated} contains $489$ abdomen CT volumes (publicly released version), featuring three segmentation targets (kidney, renal tumors and renal cyst). BraTS2023 \cite{kazerooni2023brain,menze2014multimodal} contains a total of $1,251$ 3D brain MRI volumes, each including four modalities (namely T1, T1Gd, T2, T2-FLAIR) and three segmentation targets (WT: Whole tumor, ET: Enhancing tumor, TC: Tumor core).

\subsection{Implementation Details} \label{subsec:implementation}
We use PyTorch 2.3.1-CUDA12.1 and Monai 0.9.0 to implement our experimental framework as well as the baseline models. For \textbf{FLARE2021} and \textbf{KiTS2023}, our training scheme closely follows \cite{lee20223d}, using volume patches of size $96\times96\times96$. Each input is resampled to $1.0\times1.0\times1.2 mm^3$ spacing and the models are trained for $40,000$ iterations. For \textbf{BraTS2023}, we closely follow the training scheme provided in \cite{Xin_SegMamba_MICCAI2024}. Input samples are randomly cropped to $128\times128\times128$, and models are trained for $250K$ iterations.
For all training, We use an AdamW optimizer with a learning rate of $0.0001$, Dice and CE loss as objective functions, and report the Dice similarity coefficient (Dice score) along with $95^{th}$ percentile Hausdorff Distance (HD95) for assessing volumetric accuracy. Mean scores from 5-fold cross-validation with an 80:20 split are reported for \textbf{FLARE} and \textbf{KiTS}. During BraTS training, the test set provided by the authors of \cite{Xin_SegMamba_MICCAI2024} is used for final evaluation, while the remaining dataset is split into an 80:20 ratio. All experiments used a batch size of $2$ per GPU, with 1 A100 GPU per training for FLARE and KiTS and 4 A100 GPU for BraTS. We are thankful to the maintainers of Nautilus HyperCluster by National Research Platform for providing NVIDIA A100 GPUs.

\subsection{Comparison with SOTA Methods} \label{subsec:result}
The segmentation results for the BraTS2023 dataset are listed in Table \ref{tab:brats23}. On \textbf{BraTS2023}, our model WaveFormer achieves the highest overall 91.37\% mean Dice score, with scores of 93.71\% and 88.47\% on WT and ET, respectively, and a better HD95 for TC and ET (see Table \ref{tab:brats23}). On FLARE2021, we achieved a superior overall and organ-wise (spleen, liver, and pancreas) Dice score segmentation with significantly fewer parameters compared to state-of-the-art transformer and hybrid models as shown in Table \ref{tab:flare_kits}. On KiTS2023, our model shows better overall mean Dice score (background included). KiTS2023 is a large dataset with extremely high-resolution CT scans. Due to resource constraints, we could not run different variants of our network. We plan to provide detailed results on KiTS in future work.  

\begin{table*}[!t]
    \centering
    \caption{Quantitative comparison on the BraTS2023 dataset. Best scores in bold. SegMamba parameter count was unavailable from public sources.} 

    \label{tab:brats23}
    \resizebox{1\textwidth}{!}{
    \begin{tabular} {c | c | c c c c c c c c}
    \toprule

     \multirow{3}{*}{Methods} & \multirow{3}{*}{Params$\downarrow$} & \multicolumn{8}{c}{\textbf{BraTS2023}} \\
     & & \multicolumn{2}{c}{WT} & \multicolumn{2}{c}{TC}  & \multicolumn{2}{c}{ET}  & \multicolumn{2}{c}{Avg} \\
    
    & & Dice $\uparrow$ & HD95 $\downarrow$ & Dice $\uparrow$ & HD95 $\downarrow$ & Dice $\uparrow$ & HD95 $\downarrow$ & Dice $\uparrow$ & HD95 $\downarrow$ \\
    \midrule
   
    UX-Net~\cite{lee20223d} & 53M & 93.13 & 4.56 & 90.03 &5.68 &	85.91 & 4.19  & 	89.69  & 4.81 \\
    MedNeXt~\cite{roy2023mednext} & 18M & 92.41  &4.98 & 	87.75  & 4.67  &	83.96  & 4.51  & 88.04 & 4.72 \\
    \midrule

    UNETR~\cite{hatamizadeh2022unetr} & 92.8M & 92.19 & 6.17 & 	86.39  & 5.29  & 84.48  & 5.03 & 87.68  & 5.49 \\
    SwinUNETR~\cite{hatamizadeh2022swin} & 62.2M & 92.71  & 5.22 & 	87.79  & 4.42  & 	84.21  & 4.48 & 	88.23 & 4.70 \\
    SwinUNETR-V2~\cite{he2023swinunetr} & 72.8M & 93.35 & 5.01 & 	89.65  & 4.41 & 	85.17 & 4.41  & 89.39 & 4.51 \\

    \midrule
     
    SegMamba~\cite{Xin_SegMamba_MICCAI2024} & -- & 93.61 & \textbf{3.37 } & \textbf{92.65 } & 3.85 & 87.71 & 3.48 & 91.32 & 3.56 \\

    \midrule
    \rowcolor{gray!15} 
    Our Method & \textbf{16.97M} & \textbf{93.71} & 3.64 & 91.94 & \textbf{3.67} & \textbf{88.47} & \textbf{3.26} & \textbf{91.37} & \textbf{3.52} \\

    \bottomrule
    \end{tabular}
    }
\end{table*}
\begin{table*}
    \centering
    \caption{Dice score comparison on the FLARE2021 and KiTS2023 datasets. Evaluated with background inclusion. Best scores in bold.}
    \label{tab:flare_kits}
    \resizebox{1\textwidth}{!}
    {
        \begin{tabular} {c|c| c c c c c | c}
        \toprule
        \multirow{2}{*}{Methods} & \multirow{2}{*}{Params$\downarrow$} & \multicolumn{5}{c|}{\textbf{FLARE2021}} & {\textbf{KiTS2023}} \\
         &  & Spleen & Kidney & Liver & Pancreas & Mean$\uparrow$ & Mean$\uparrow$\\
        \midrule
       
        TransBTS~\cite{wenxuan2021transbts} & 31.6M & 96.4 & 95.9 & 97.4 & 71.1 & 90.2 & 75.56 \\
        UNETR~\cite{hatamizadeh2022unetr} & 92.8M & 92.7 & 94.7 & 96.0 & 71.0 & 88.6 & 70.23 \\
        \midrule
        SwinUNETR~\cite{hatamizadeh2022swin} & 62.2M & 97.9 & 96.5 & 98.0 & 78.8 & 92.9 & 80.74 \\
        nnFormer~\cite{zhou2021nnformer} & 149.3M & 96.0 & \textbf{97.5} & 97.7 & 71.7 & 90.8 & 78.51 \\
        3D-UXNET~\cite{lee20223d} & 53M & 98.1 & 96.9 & \textbf{98.2} & 80.1 & 93.4 & 78.88 \\
        \midrule
        \rowcolor{gray!15} 
        Ours & \textbf{16.9M} & \textbf{98.2} & 97.0 & \textbf{98.2} & \textbf{81.7} & \textbf{93.8} & \textbf{80.91} \\    
        \bottomrule
        \end{tabular}
    }
\end{table*}



   


\subsection{Ablation of Architectural Improvements} We experiment with the capability of our network on multi-scale tumor detection shown in Table \ref{tab:ablation}. Multi-level attention does increase the segmentation capability for small-size TC and ET with its local window context learning capability. HF-refinement increases the performance slightly for small-scale but sometimes hurts the performance of large/medium-scale tumors. We believe improved multi-scale organ detection is possible with a better HF-Refinement technique. Attention weights have been shared across multi-level attention which is a space for future exploration.

\begin{table*}[!t]
    \centering
    \caption{Dice score comparison on BraTS across scales and architectural variations. Binning for Small(S), Medium(M), Large(L) targets (in $cm^3$): WT: [0–71, 71–120, >120], TC: [0–20, 20–40, >40], ET: [0–12, 12–30, >30]}
    \label{tab:ablation}
    \resizebox{1\textwidth}{!}{%
    \begin{tabular}{c|c|c|c|c|c|c|c|c|c|c|c}
    \toprule
    \multirow{2}{*}{Methods} 
    & \multirow{2}{*}{Params} 
    & \multicolumn{3}{c|}{WT (\%)} 
    & \multicolumn{3}{c|}{TC (\%)} 
    & \multicolumn{3}{c|}{ET (\%)} 
    & \multirow{2}{*}{Mean} \\
    \cmidrule(lr){3-11}
     & & S & M & L 
     & S & M & L 
     & S & M & L & \\
    \midrule
    \makecell{Simple Up \\(2-Transpose Conv)}
     & 19.3M
     & 90.61 & 93.99 & 94.92
     & 84.87 & 92.85 & 93.70
     & 79.34 & 91.37 & 91.92 & 90.20 \\
    \midrule
    \makecell{Residual Up}
     & 16.97M
     & 91.41 & 94.58 & 95.24
     & 85.83 & 94.75 & 94.70
     & 81.26 & 92.72 & 93.17 & 91.12 \\
    \midrule
    \makecell{Residual Up + \\ HF Ref.}
     & 17.06M
     & 91.44 & 94.67 & 95.26
     & 85.88 & 93.98 & 94.77
     & 81.51 & 92.72 & 92.71 & 90.97 \\
    \midrule
    \makecell{Residual Up + \\ Multi-level Attn}
     & 16.97M
     & 91.04 & 94.67 & 95.40
     & 87.90 & 94.74 & 94.96
     & 82.13 & 92.64 & 93.07 & 91.37 \\
    \bottomrule
    \end{tabular}}
\end{table*}

\section{Conclusion}
In this work, we introduced WaveFormer, a novel 3D Transformer model leveraging discrete wavelet transformations for efficient medical image segmentation. Our approach effectively captures both global context and fine-grained details while significantly reducing computational overhead and parameter count. Experimental results demonstrate that WaveFormer outperforms state-of-the-art models in certain organ segmentation tasks with smaller model size, and achieves comparable performance on others, maintaining efficiency without compromising accuracy. Our work underscores the potential of frequency-domain representations in advancing lightweight and effective deep learning solutions for medical image analysis.
\bibliographystyle{splncs04}
\bibliography{references}

\end{document}